\documentclass{svproc}
\usepackage[english]{babel}
\usepackage{booktabs}
\usepackage{mathtools}
\usepackage{makecell}
\usepackage{graphicx}
\usepackage{algorithm}
\usepackage{algorithmic}
\usepackage[export]{adjustbox}
\usepackage{eurosym} 
\usepackage{color}
\usepackage{float}
\usepackage{cite}
\graphicspath{{figures/}}
\usepackage{url}

\usepackage{float}
\usepackage[caption = false]{subfig}
\DeclareMathOperator{\atan2}{atan2}

\begin{document}
\mainmatter            

\title{Human Intention Recognition in Flexible Robotized Warehouses based on Markov Decision Processes}
\titlerunning{Human Intention Recognition}

\author{Tomislav Petkovi\'{c} \and Ivan Markovi\'{c} \and Ivan Petrovi\'{c}}
\authorrunning{Tomislav Petkovi\'{c} et al.}
\tocauthor{Tomislav Petkovi\'{c}, Ivan Markovi\'{c}, Ivan Petrov\'{c}}

\institute{University of Zagreb Faculty of Electrical Engineering and Computing, Croatia\\
\email{\{tomislav.petkovic2, ivan.markovic, ivan.petrovic\}@fer.hr}}
\maketitle           

\begin{abstract}
The rapid growth of e-commerce increases the need for larger warehouses and their automation, thus using robots as
assistants to human workers becomes a priority.
In order to operate efficiently and safely, robot assistants or the supervising system should recognize human
intentions. 
Theory of mind (ToM) is an intuitive conception of other agents' mental state, i.e., beliefs and desires, and how they
cause behavior. 
In this paper we present a ToM-based algorithm for human intention recognition in flexible robotized warehouses.
We have placed the warehouse worker in a simulated 2D environment with three potential goals.
We observe agent's actions and validate them with respect to the goal locations using a Markov decision process
framework.
Those observations are then processed by the proposed hidden Markov model framework which estimated agent's desires.
We demonstrate that the proposed framework predicts human warehouse worker's desires in an intuitive manner and in the
end we discuss the simulation results.
\end{abstract}

\section{Introduction}
The European e-commerce turnover managed to increase $13.3\%$ to \EUR{455.3} billion in 2015, compared to the $1.0\%$
growth of general retail in Europe \cite{E-commerceEurope2016}. With the internationalization of distribution chains,
the key for success lies within efficient logistics, consequently increasing the need for larger warehouses and their
automation. There are many fully automated warehouse systems such as the Swisslog's CarryPick Mobile system and Amazon's
Kiva system \cite{d2012guest}. 
They use movable racks that can be lifted by small, autonomous robots. By bringing the product to the worker,
productivity is increased by a factor of two or more, while simultaneously improving accountability and flexibility
\cite{Wurman2008}. However, current automation solutions based on strict separation of humans and robots provide limited
operation efficiency of large warehouses.
Therefore, a new integrated paradigm arises where humans and robots will work closely together and these integrated
warehouse models will fundamentally change the way we use mobile robots in modern warehouses.
Besides immediate safety issues, example of a challenge such models face, is to estimate worker's 
intentions so that the worker may be assisted and not impeded in his work.
Furthermore, if the robot is not intelligent, but controlled by a supervisory system, the supervisory system needs to be
able to estimate worker's intentions correctly and control the robots accordingly, so that the warehouse operation
efficiency is ensured.

There exists a plethora of challenges in human intention recognition, because of the subtlety and diversity of human
behaviors \cite{Bandyopadhyay2013}. 
Contrary to some more common quantities, such as the position and velocity, the human intention is not directly
observable and needs to be estimated from human actions.
Furthermore, the intentions should be estimated in real-time and overly complicated models should be avoided.
Having that in mind, only the actions with the greatest influence on intention perception should be considered.
For example, in the warehouse domain, worker's orientation and motion have large effect on the goal intention
recognition.
On the other hand, observing, e.g., worker's heart rate or perspiration could provide very few, if any, information on
worker's intentions. 
Therefore, such measurements should be avoided in order to reduce model complexity and ensure real-time operation
\cite{Bandyopadhyay2013}.

Many models addressing the problem of human intention recognition successfully emulate human social intelligence using
Markov decision processes (MDPs).
The examples of such models can be found in \cite{Bandyopadhyay2013}, where authors propose framework for estimating
pedestrian's intention to cross the road and in \cite{Lin2014} where authors proposed framework for gesture recognition
and robot assisted coffee serving.
There are multiple papers from the gaming industry perspective, proposing methods for improving the non-playable
character's assisting efficiency \cite{Nguyen2012,Fern2011}. 
An interesting approach is the Bayesian Theory of Mind (BToM) \cite{Baker2014a} where beliefs and desires generate
actions via abstract causal laws.
BToM framework in \cite{Baker2014a} observes agent's actions and estimates agent's desires to eat at a particular
food-truck.
However, though impressive, BToM model does not predict the possibility of agent's change of mind during the simulation
\cite{Chater2010}. 

In this paper, we propose an algorithm for warehouse worker intention recognition motivated by the BToM approach.
We expand the BToM to accommodate the warehouse scenario problem and we present formal and mathematical details of the
constructed MDP framework.
The warehouse worker is placed in a simulated 2D warehouse environment with multiple potential goals, i.e., warehouse
items that need to be picked.
The worker's actions, moving and turning, are validated based on the proposed MDP framework.
Actions, resulting in motion towards the goal, yield greater values than those resulting in moving away from the goal.
We introduce worker's intention recognition algorithm based on the hidden Markov model (HMM) framework similarly to
those presented in \cite{Wang2009,He2012}.
The proposed intention recognition algorithm observes action values generated by the MDP and estimates potential goal
desires.
We have considered worker's tendency to change its mind during the simulation, as well as worker's intention of leaving
the rack.
In the end, we demonstrate that the proposed algorithm predicts human warehouse worker's desires in an intuitive manner and we discuss the simulation results.

\section{Human Action Validation} \label{validation}
\begin{figure}[htb]
\vspace{-3.8cm}
\begin{minipage}[!t]{0.45\textwidth}
	\includegraphics[width=\textwidth]{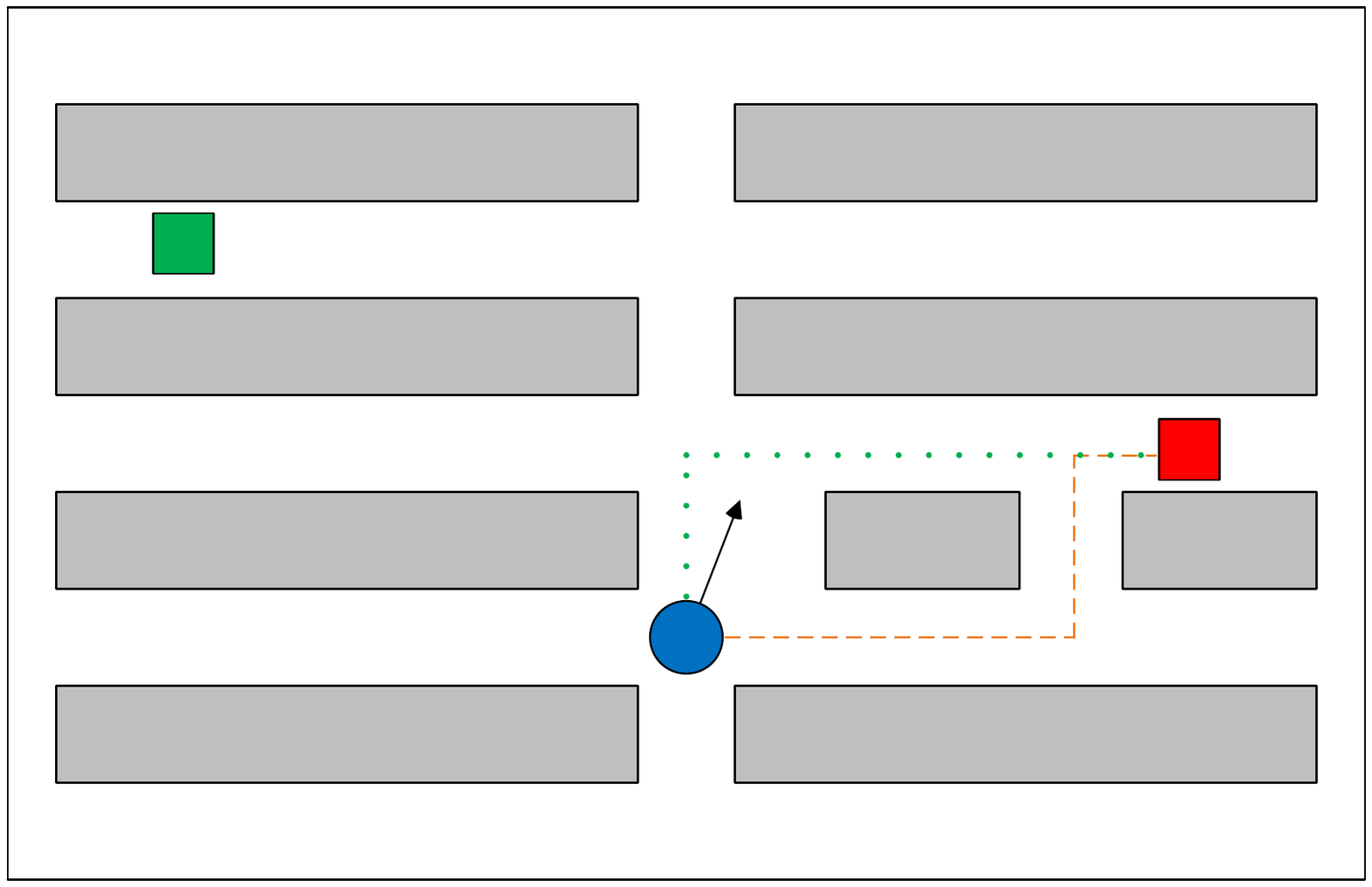}
	\caption{Warehouse worker's (blue circle) shortest path to the red goal is ambiguous because both orange dashed and green dotted paths are optimal. The black arrow denotes worker's orientation.}
	\label{fig:astar_example}
\end{minipage}
\qquad
\begin{minipage}[!t]{0.5\textwidth}
\vspace{2.9cm}

    \subfloat[]{\includegraphics[width= .44\textwidth]{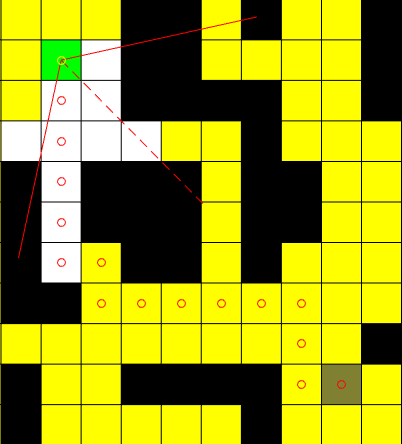}}			\qquad
    \subfloat[]{\includegraphics[width= .44\textwidth]{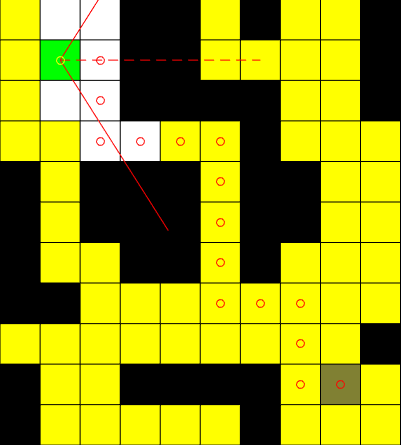}}   
    \label{fig:astar_right}
    \caption{The proposed A$^*$ modification yields different optimal paths with the agent's orientation change.}
	\label{fig:astar}
\end{minipage}
\hspace{-2cm}
\end{figure}

\begin{figure}[htb]
\centering
\includegraphics[width=.6\textwidth]{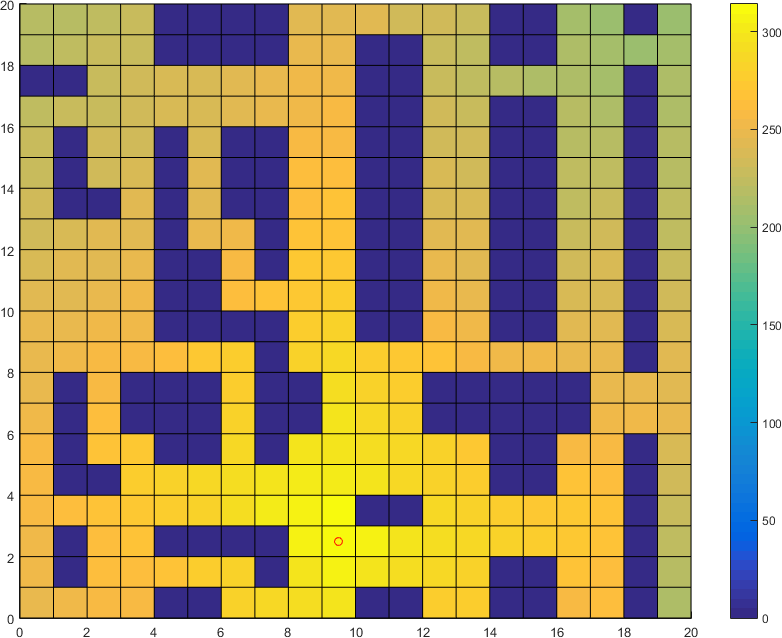}
\caption{State values for the agent's orientation of $\frac{3\pi}{2}$ if the goal state is the southern goal labeled with the red circle.}
\label{fig:viter_figure}
\vspace{-0.5cm}
\end{figure}

In the integrated warehouse environment the worker's position and orientation need to be precisely estimated, and in the
present paper we assume that these quantities are readily available.
Furthermore, we would like to emphasize that most of warehouse worker duties, such as sorting
and placing materials or items on racks, unpacking and packing, include a lot of motion which is
considered to be inherently of the stochastic nature.
Therefore, we model the worker's perception model $P(O|S)$, with $O$ and $S$ representing observation and state,
respectively, as deterministic, and the action model $P(S'|S, A)$, with $A$ representing action, as stochastic.
A paradigm that encompasses uncertainty in agent's motion and is suitable for the problem at hand are MDPs
\cite{Thrun1999}.
The MDP framework is based on the idea that transiting to state $S$ yields an immediate
reward $R$. Desirable states, such as warehouse items worker needs to pick up, have high immediate reward values, while undesirable
states, such as warehouse parts that are of no immediate interest, have low immediate reward values.
The rational worker will always take actions that will lead to the highest total expected reward and that value needs to
be calculated for each state.
\begin{figure}[!b]
\centering
\includegraphics[width=.6\textwidth]{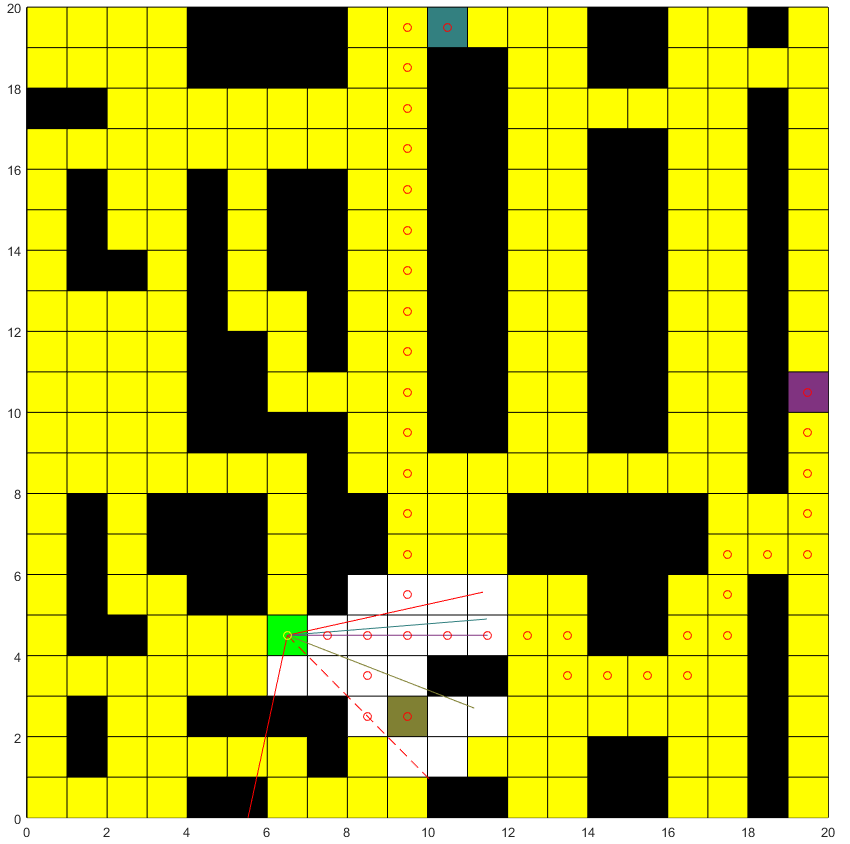}
\caption{Agent (green tile) in simulation environment with three potential agent's goals (colored tiles). Unoccupied
space is labeled with yellow tiles and occupied space (i.e. warehouse racks) is labeled with black tiles. The optimal
path to each goal is shown with red dots and red solid lines denote agent's vision field. Visible tiles are colored white. Red dashed line denotes agent's current orientation and colored lines indicate direction of the average orientation of the visible
optimal path to each goal calculated using \eqref{eq:avg_or}.}
\label{fig:human_warehouse}
\vspace{-0.5cm}
\end{figure}
Before approaching that problem, we define the MDP framework applicable to the warehouse domain.
In order to accomplish that, we have placed the worker (later referred as agent) in a simulated 2D warehouse environment
shown in Fig.~\ref{fig:human_warehouse}. The environment is constructed using MATLAB\textsuperscript{\textregistered} GUI development environment without predefined physical interpretation of the map tiles and the map size is chosen arbitrary to be $20\times20$. There are three potential goals and the shortest path to each goal is calculated. There are many off-the-shelf graph search algorithms we can use to find the optimal path to each goal, such as
Dijkstra's algorithm and A$^*$. 
However, if there exist multiple optimal paths to the goal, there are no predefined rules which one to select and
the selection depends on the implementation details.
Consider the example scenario with the warehouse worker in Fig. \ref{fig:astar_example}.
It is intuitive that the rational worker will tend to follow the green path because the orange path would require the
worker to either take the additional action of turning or unnatural walking by not looking forward.
Having this in mind, we modify the A$^*$ search algorithm which selects optimal path the
agent currently sees the most.
 This has been done by introducing the heuristic matrix $H$ using the Manhattan distance ($L_1$) heuristics as follows:
\begin{equation}
H_{x,y}= 
\begin{dcases}
    \ |x_g-x|+|y_g-y|-\epsilon, & \text{if the agent sees tile (x,y)} \\
    \ |x_g-x|+|y_g-y|, & \text{otherwise}
\end{dcases}
\label{eq:Hmatrix}
\end{equation}
where $\epsilon$ is a small value. Subtracting a small value from the visible tiles directs the search in their direction and does not disrupt heuristic's admissibility. The cost of each movement is also modified in a
similar way by subtracting a small value $\epsilon$ from the base movement cost of $1$, if the tile is visible.
Example of the modified A$^*$ search algorithm results can be seen in Fig.~\ref{fig:astar}.
The average orientation of the visible path to each goal is defined as follows:
\begin{equation}
\theta_{goal}=
\begin{dcases}
    \atan2\Big(\sum_{n=1}^{N}\sin(\theta_n), \sum_{n=1}^{N}\cos(\theta_n)\Big) ,& \text{if } N>0\\
    \theta_a+\pi,              & \text{otherwise}
\end{dcases}
\label{eq:avg_or}
\end{equation}
where $N$ is the number of visible optimal path tiles, $\theta_a$ is agent's orientation and $\theta_n$ are relative
orientations of each visible optimal path tile ($x$, $y$) with respect to the agent ($x_a$, $y_a$):
\begin{equation}
\theta_{n}= \atan2(y-y_a, x-x_a).
\label{eq:theta_n}
\end{equation}
We propose a mathematical model for validating agent's actions based on the assumption that the rational agent tends to (i)
move towards the goal it desires most by taking the shortest possible path, and (ii) orients in a way to minimize difference
between its orientation and most desirable goal's average orientation of the visible optimal
path calculated in \eqref{eq:avg_or}. 
The proposed model goal is to assign large value to the actions compatible with the mentioned assumptions, and small
values to the actions deviating from them. 
These values will be used to develop agent's intention recognition algorithm in the sequel.
We can notice that the introduced validation problem is actually a path planning optimization
problem. 
Perfectly rational agent will always choose the action with the greatest value and consequently move towards
the goal. 
We approach the agent's action values calculation by introducing the agent's action validation MDP framework.
We assume that agent's position and orientation are fully observable and create the MDP state space $S$ as:
\begin{equation}
S_{x,y,k}=\begin{bmatrix} x\\ y\\ \theta
\end{bmatrix}.
\end{equation}
The agent's orientation space $\Theta$ must be discrete because the MDP framework assumes a finite number of states. We
have modeled $\Theta$ to include orientations divisible with $\frac{\pi}{4}$ and it can be arbitrary expanded:
\begin{equation}
\Theta=\{0, \frac{\pi}{4},  \frac{\pi}{2},  \frac{3\pi}{4}, \pi,  \frac{5\pi}{4},  \frac{3\pi}{2},  \frac{7\pi}{4} \}.
\end{equation}
The action space $A$ includes actions  `Up', `Down', `Left', `Right', `Turn Clockwise', `Turn Counterclockwise' and `Stay', labeled in order as follows: 
\begin{equation}
A=(\wedge,\vee,  <, >, R, L, S).
\label{eq:actions}
\end{equation}
It has already been stated that the agent's actions are fully observable but stochastic.
In order to capture stochastic nature of the agent's movement, we define the transition matrix $T$ of agent's
movement: 
\begingroup
\renewcommand*{\arraycolsep}{5pt}
\begin{equation}
T=\begin{bmatrix}
1-2\epsilon & 0 & \epsilon & \epsilon & 0 & 0 & 0 \\      
0 & 1-2\epsilon & \epsilon & \epsilon & 0 & 0 & 0 \\
 \epsilon &  \epsilon & 1-2\epsilon & 0 & 0 & 0 & 0 \\
 \epsilon &  \epsilon & 0 & 1-2\epsilon & 0 & 0 & 0 \\
0 & 0 & 0 & 0 & 1-\epsilon & 0 &  \epsilon \\
0 & 0 & 0 & 0 & 0 & 1-\epsilon & \epsilon \\
0 & 0 & 0 & 0 & 0 & 0 & 1 \\
\end{bmatrix}
\label{eq:transition}
\end{equation}
\endgroup
where element $T_{ij}$ denotes realization probability of the action $A_j$, if the wanted action is $A_i$. Moving
actions have small probability $2\epsilon$ of resulting in lateral movement, and turning actions have small probability $\epsilon$ of failing. The value of the constant $\epsilon$ is obtained experimentally and equals to $0.1$.
If the agent's action cannot be completed, because of the occupied space blocking the way, column responding to the
impossible action is added to last column and is set to zero vector afterwards. We define three hypotheses, $H_i, i=1\dots 3$, one
for each goal state as follows: \textit{``Agent wants to go to the goal $i$ and other potential goals are treated as
unoccupied tiles''}. The immediate reward values $R$ for each hypothesis and state are calculated as follows:
\begin{equation}
R_{i, S'}= 
\begin{dcases}
    \pi,& \text{if \textit{S'} is the goal state according to the } H_i	\\
    -(\epsilon+|\theta_i-\theta_a|),              & \text{otherwise}
\end{dcases}
\label{eq:R_calc}
\end{equation}
where $\epsilon$ is a small number and $|\theta_i-\theta_a|$ represents the absolute difference between average
orientation of the visible path to the goal \textit{i} and agent's orientation. Note that we have taken the angle
periodicity into account while calculating the angle difference in \eqref{eq:R_calc}. The goal state is rewarded and
other states are punished proportionally to the orientation difference. If the agent does not see path to the goal
\textit{i}, the reward is set to the lowest value, $-\pi$ which is derived from \eqref{eq:theta_n}. One of the most
commonly used algorithms for solving the MPD optimal policy problem is the value iteration algorithm
\cite{Bandyopadhyay2013}, which assigns calculated value to the each state. The optimal policy is derived by choosing the
actions with the largest expected value gain. The value iteration algorithm iteratively solves the Bellman's equation
\cite{Bellman1957a} for each hypothesis $H_i$:
\begin{equation}
V_{j+1}(H_i,S)=\underset{a}{\max}\{\sum_{S'}P_{S,S'}(R_{H_i,S'}+\gamma V_j(H_i, S') )\}
\label{eq:bellman}
\end{equation}
where $S$ is the current state, $S'$ adjacent state, and $P_{S,S'}$ element of the row $T_a$ in transition matrix $T$ which would cause transitioning from state $S$ to $S'$. The algorithm stops once the criteria:
\begin{equation}
\sum_{i,k}||V_{j}(H_i, S_k)-V_{j-1}(H_i, S_k)||<\eta
\end{equation}
is met, where the threshold $\eta$ is set to $0.01$. State values, if the goal state is the dark yellow (southern) goal and agent's orientation of $\frac{3\pi}{2}$, is shown in Fig.~\ref{fig:viter_figure}.
The agent's behavior consistent with the hypothesis $H_i$ is defined as follows.
\begin{definition}{(Consistent behavior)} If the agent in state $S$ takes the action $a$ under the hypothesis $H_i$, with the expected value gain greater or equal than the expected value gain of the action ``Stay'', its behavior is considered consistent with the hypothesis $H_i$. Otherwise, its behavior is considered inconsistent with the hypothesis $H_i$.
\end{definition}
Behavior consistency is an important factor in determining agent's rationality, which will be further discussed in next
section. While calculating the immediate rewards and state values has $\mathcal{O}(n^4)$ complexity and can be time
consuming, it can be done offline, before the simulation start. Optimal action, $\Pi^*(H_i,S)$, for each state is the
action that maximizes expected value gain and, on the other hand, the worst possible action, $\bar{\Pi}^*(H_i,S)$, is
the action that minimizes expected value gain.

\section{Human Intention Recognition} \label{intention}

\begin{figure}[!t]
\centering
\includegraphics[width=.7\textwidth]{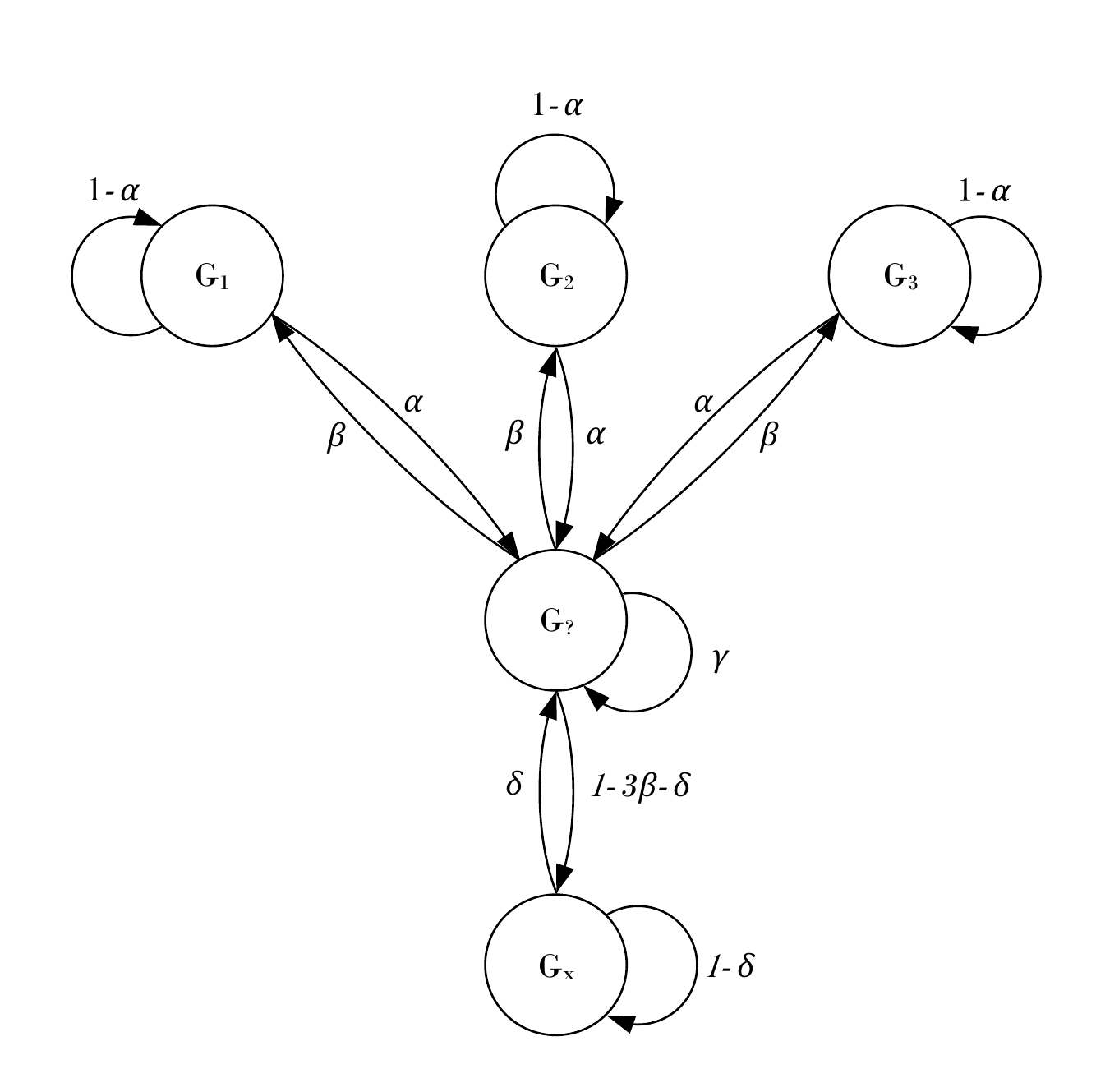}
\caption{Hidden states and transition probabilities. The used constant values are as follows: $\alpha=0.2$, $\beta=0.1$, $\gamma=0.65$, $\delta=0.1$.}
\label{fig:HMM_states}
\vspace{-0.6cm}
\end{figure}

Once the state values $V_{H,S}$ are obtained, model for solving agent's intention recognition is introduced. While
agent's actions are fully observable, they depend on agent's inner states (desires), which cannot be observed and need
to be estimated. We propose framework based on hidden Markov model for solving the agent's desires estimation
problem. HMMs are especially known for their application in temporal pattern recognition such as speech, handwriting,
gesture recognition \cite{Rabiner1989a} and force analysis \cite{Chen1997}. They are an MDP extension including the
case where the observation (agent's action) is a probabilistic function of the hidden state (agent's desires) which
cannot be directly observed. We propose a model with five hidden states, which is shown in Fig.~\ref{fig:HMM_states} and
listed in Table~\ref{tbl:HMM_states}.
\begin{table}[!t]
\caption{HMM framework components}
\label{tbl:HMM_states}
\centering
\begin{tabular}{lll} \toprule
Symbol & Name & Description\\ \midrule
$G_1$ & Goal 1 & Agent wants to go to the northern goal \\
$G_2$ & Goal 2 & Agent wants to go to the eastern goal \\
$G_3$ & Goal 3 & Agent wants to go to the southern goal \\
$G_?$ & Unknown goal & Agent's desires are not certain \\
$G_x$ & Irrational agent & Agent behaves irrationally \\
\end{tabular}
\vspace{-0.9cm}
\end{table}
In order to avoid confusion caused by MDP and HMM frameworks, both having similar or the same element names, MDP states
will be referred as states and HMM states will be referred to as hidden states, or simply desires. All of
the other similarly named elements in this chapter refer to the HMM unless stated otherwise. Hidden state $G_?$
indicates that the agent behaves consistent with multiple goal hypotheses and the model cannot decide between them with
enough certainty. On the other hand, hidden state $G_x$ indicates that the agent behaves inconsistently with every goal
hypothesis. This hidden state includes the cases of the agent being irrational or agent's desire to go to an a priori
unknown goal. The proposed model cannot distinguish between these cases. The agent's change of mind during the
simulation is allowed, but with very low probability. The constant values in Fig.~\ref{fig:HMM_states} are obtained
experimentally and we introduce HMM transition matrix $T_{HMM}$:
\begingroup
\renewcommand*{\arraycolsep}{5pt}
\begin{equation}
T_{HMM}=\begin{bmatrix}
0.8& 0 & 0 & 0.2 & 0 \\      
0& 0.8 & 0 & 0.2 & 0 \\      
0& 0 & 0.8 & 0.2 & 0 \\      
0.1& 0.1& 0.1 & 0.65 & 0.05 \\      
0& 0 & 0 & 0.1 & 0.9 \\      
\end{bmatrix}.
\label{eq:transition_HMM}
\end{equation}
\endgroup
During the simulation, each agent's action $a \in A$ generates a three element observation vector $O_{S, a}$, each
element belonging to one hypothesis. Observation vector element $O_{S, a, i}$ is calculated as follows:  
\begin{equation}
O_{S, a, i}=\frac{E[V(H_i,S)|a]-E[V(H_i,S)| \bar{\Pi}^*]}{E[V(H_i,S)| {\Pi}^*]-E[V(H_i,S)| \bar{\Pi}^*]},
\label{eq:observation}
\end{equation}
where $E[\cdot]$ denotes expected value gain. Calculated observations are used to generate the HMM emission matrix $B$. The emission matrix is expanded with each agent's action (simulation step) with the row $B'$, where the element $B'_{i}$ stores the probability of observing observation vector $O$ from hidden state $S_i$. Last three observations are averaged and maximum average value $\phi$ is selected. It is used as an indicator if the agent is behaving irrationally. Each expansion row $B'$ is calculated as
follows:
\begin{equation}
B'=\zeta \cdot \begin{dcases}
\begin{bmatrix} \tanh(O_1) & \tanh(O_2)& \tanh(O_3) & \tanh(0.55) & 0 \end{bmatrix}, & \text{if } \phi>0.5\\
\begin{bmatrix} \tanh(\frac{O_1}{2}) & \tanh(\frac{O_2}{2})& \tanh(\frac{O_3}{2}) & \tanh(0.1) & \tanh(1-\phi) \end{bmatrix}, & \text{otherwise}
\end{dcases}
\label{eq:Bmatrix}
\end{equation}
where $\zeta$ is a normalizing constant and $O$ is $i$-th observation vector. The initial probabilities of agent's
desires are:
\begin{equation}
\Pi= \begin{bmatrix}
0 & 0 & 0 & 1 & 0
\end{bmatrix},
\label{eq:initial_desires}
\end{equation}
indicating that the initial state is $G_?$. After each agent's action, the agent's desires are estimated using the
Viterbi algorithm \cite{Forney1973} which is often used for solving HMM human intention recognition models
\cite{Zhu2008}. The Viterbi algorithm outputs the most probable hidden state sequence and the probabilities of each
hidden state in each step. These probabilities are the agent's desire estimates.

\section{Simulation Results}  \label{simulation}
\begin{figure}[!h]
\vspace{-0.5cm}
\centering
\subfloat[Simulation step 2.]{\includegraphics[width= .33\textwidth]{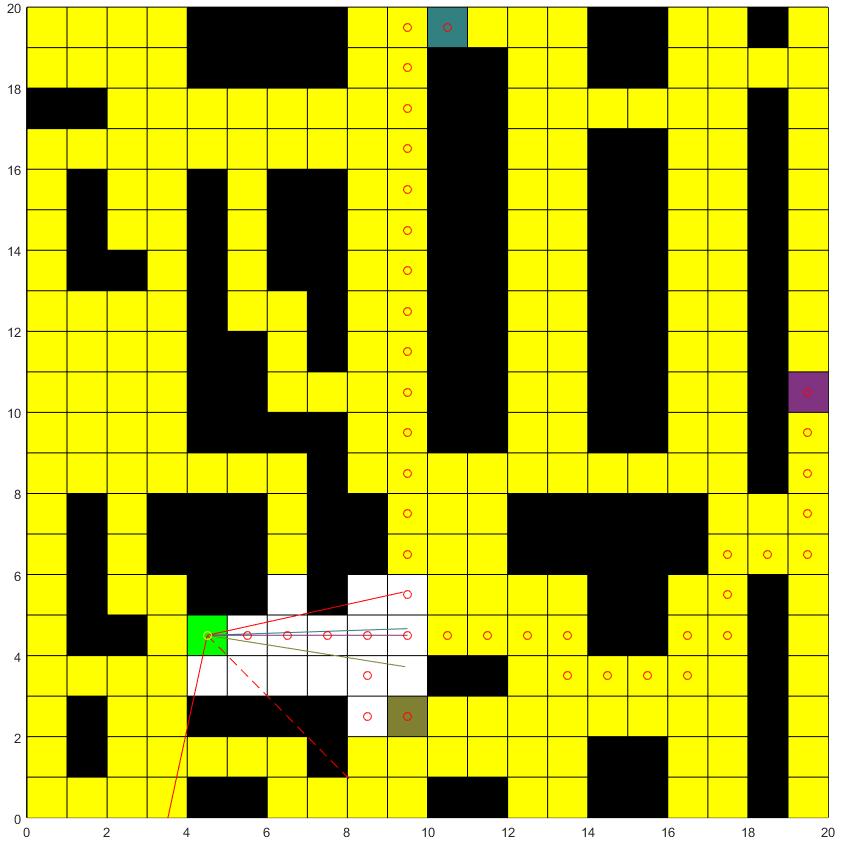}}
\subfloat[Simulation step 3.]{\includegraphics[width= .33\textwidth]{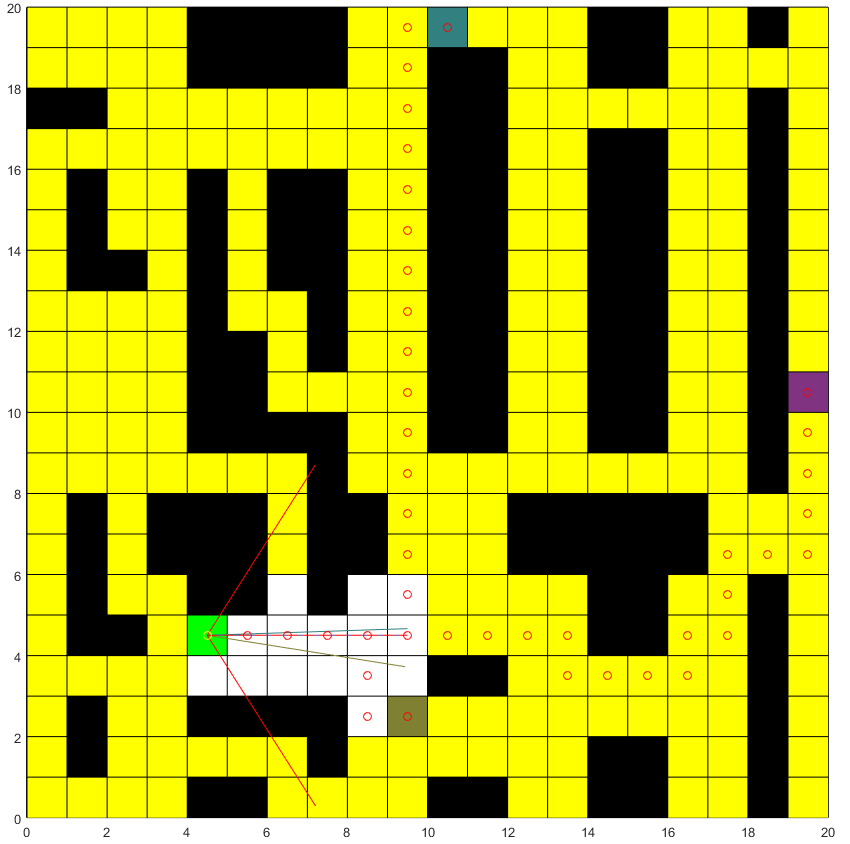}}
\subfloat[Simulation step 6.]{\includegraphics[width= .33\textwidth]{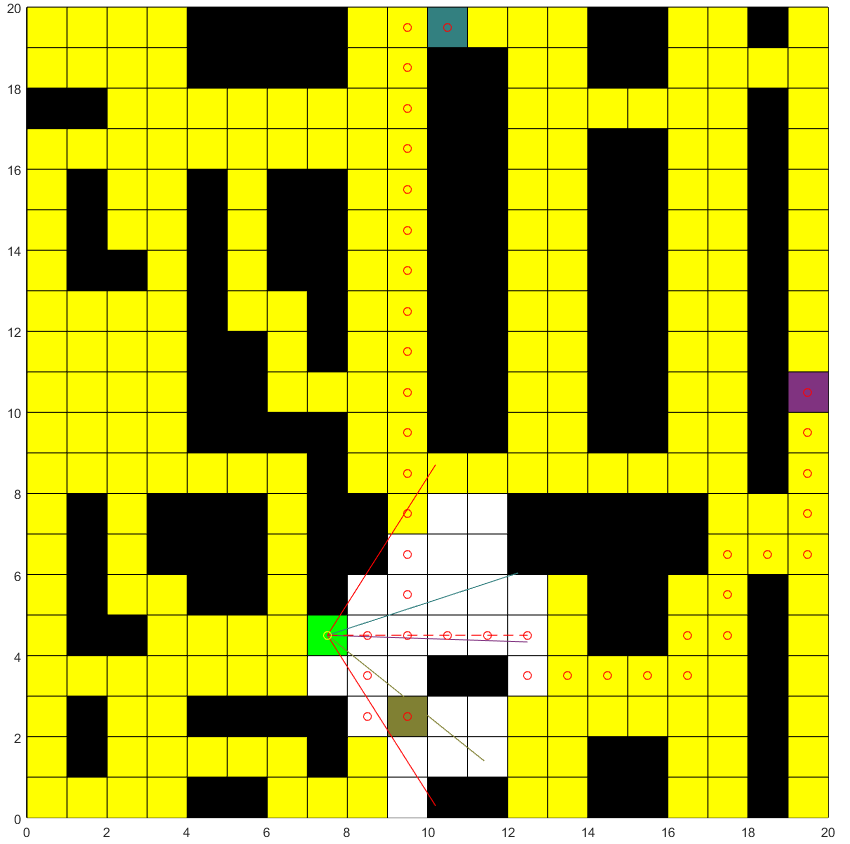}}

\subfloat[Simulation step 7.]{\includegraphics[width= .33\textwidth]{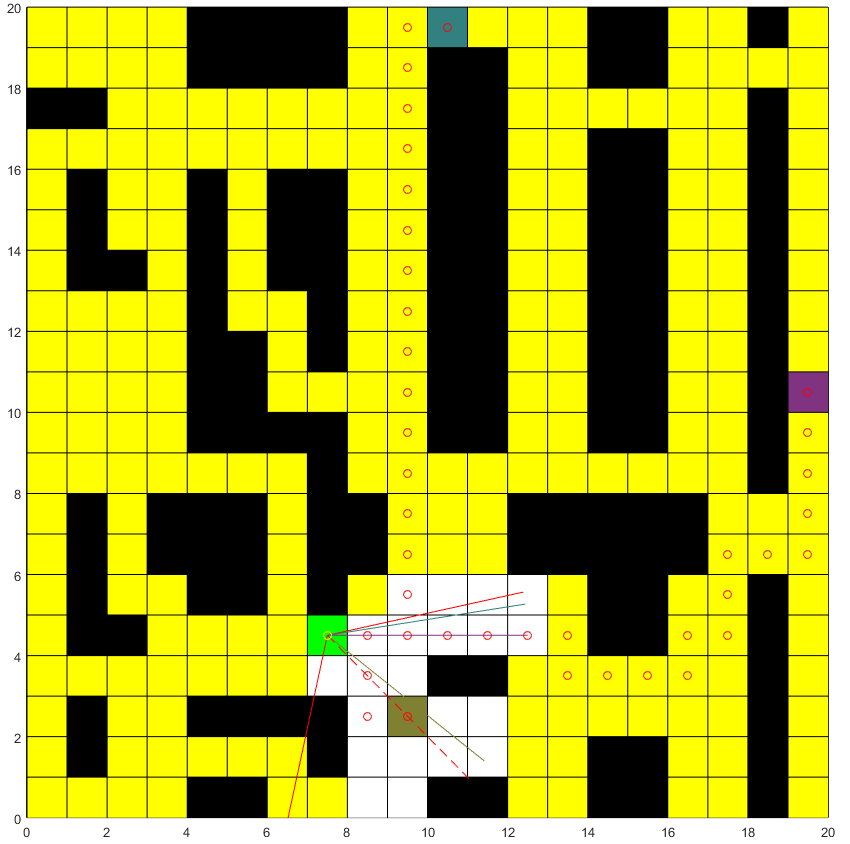}}
\subfloat[Simulation step 12.]{\includegraphics[width= .33\textwidth]{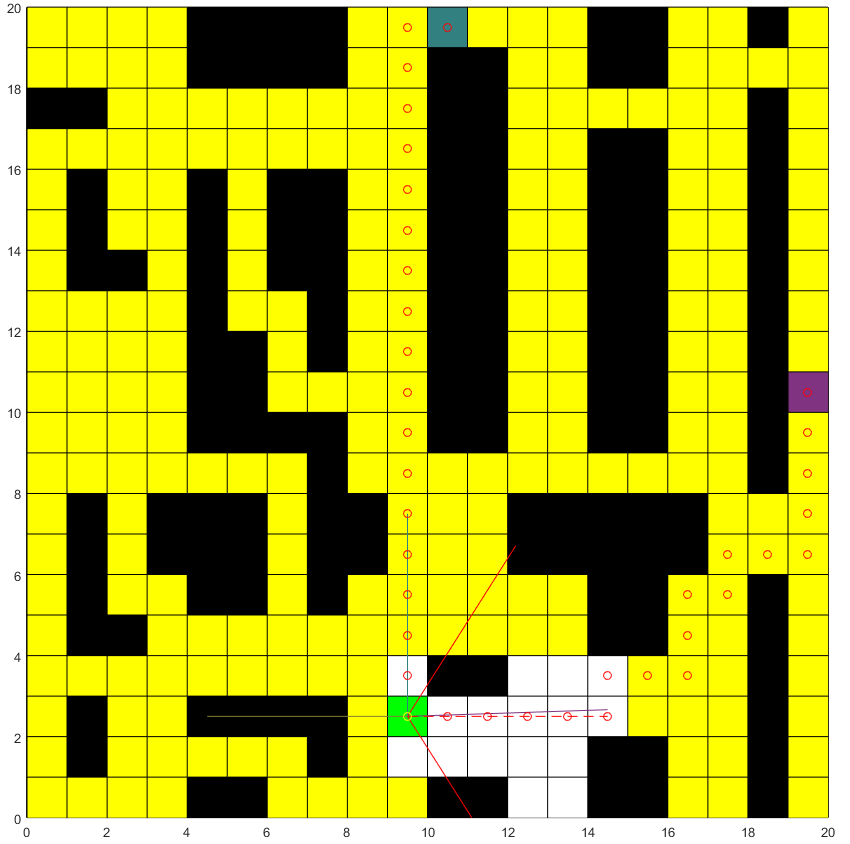}}
\subfloat[Simulation step 13.]{\includegraphics[width= .33\textwidth]{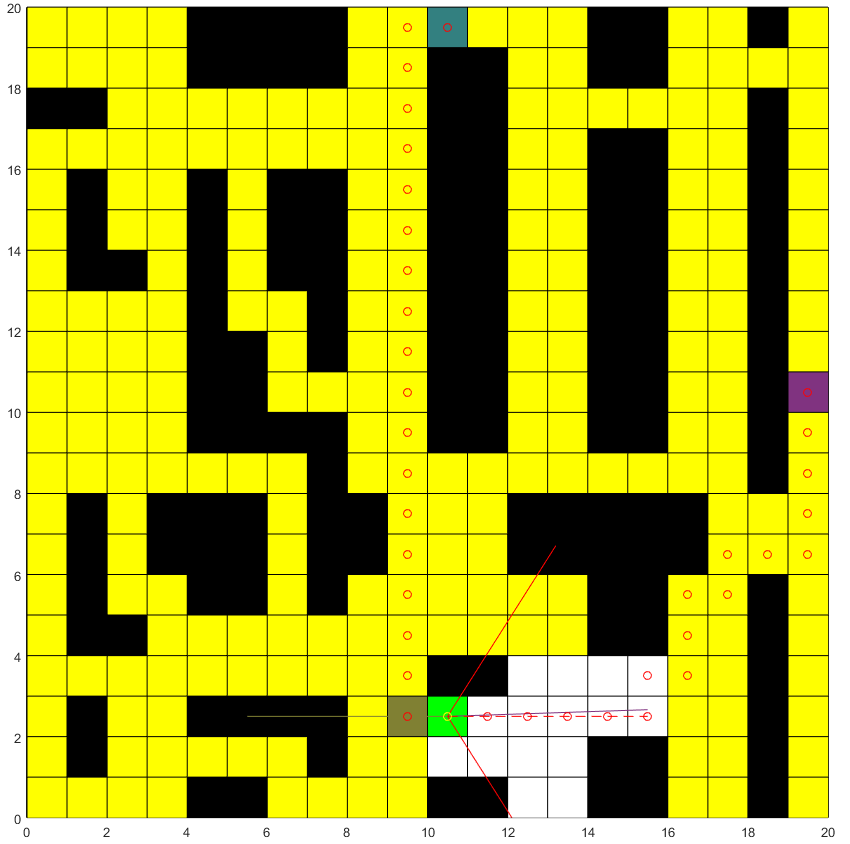}}

\subfloat[Simulation step 18.]{\includegraphics[width= .33\textwidth]{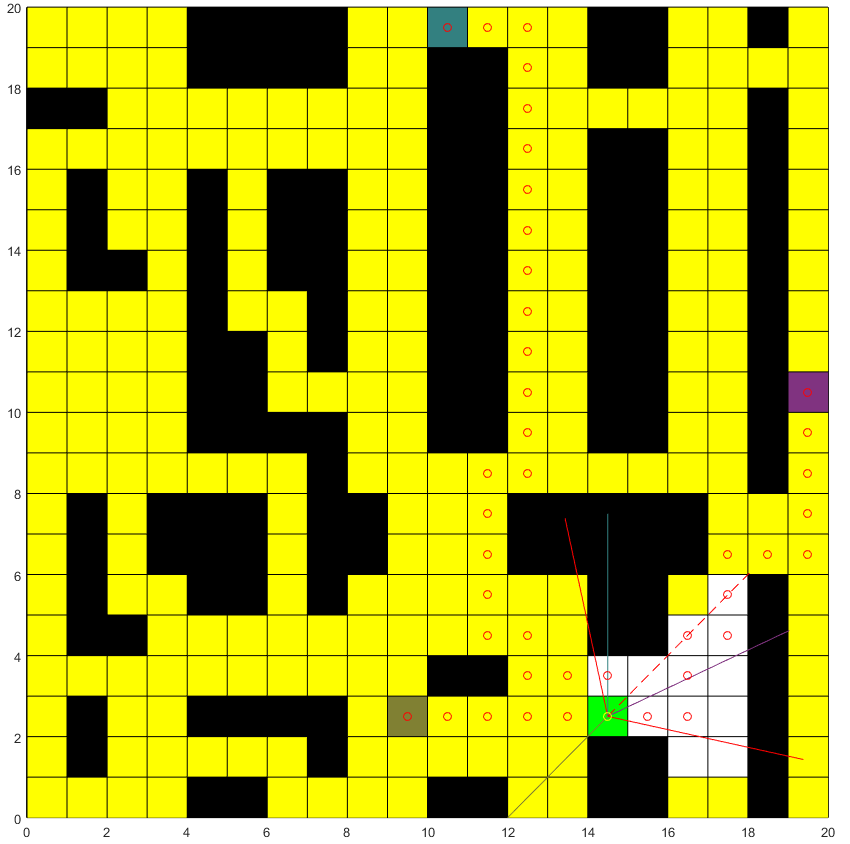}}
\subfloat[Simulation step 25.]{\includegraphics[width= .33\textwidth]{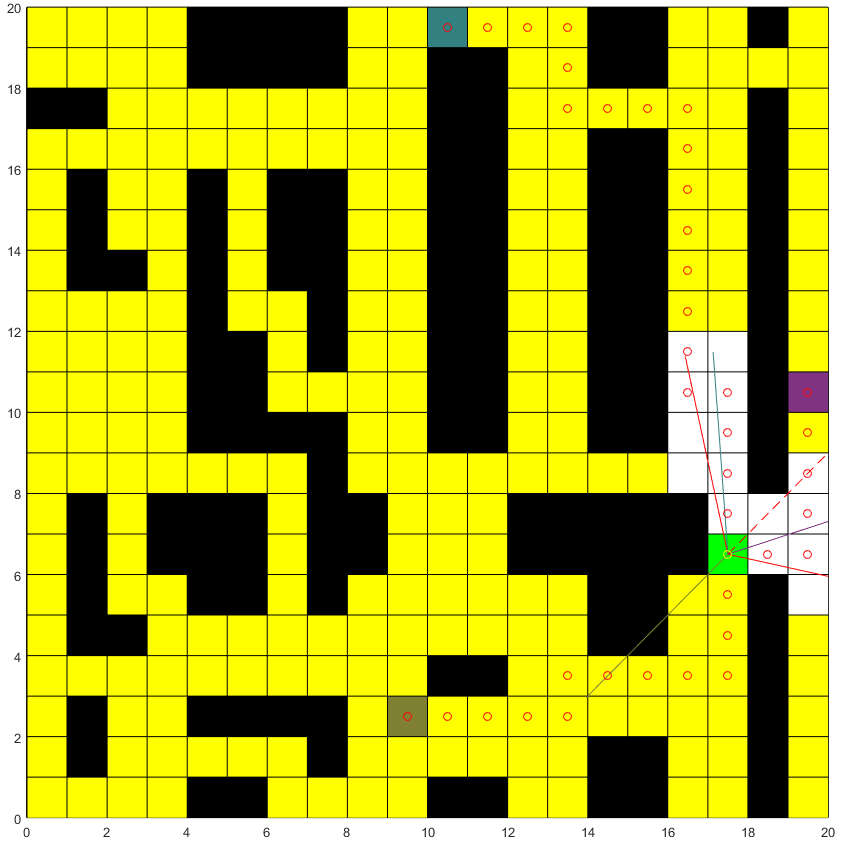}}
\subfloat[Simulation step 31.]{\includegraphics[width= .33\textwidth]{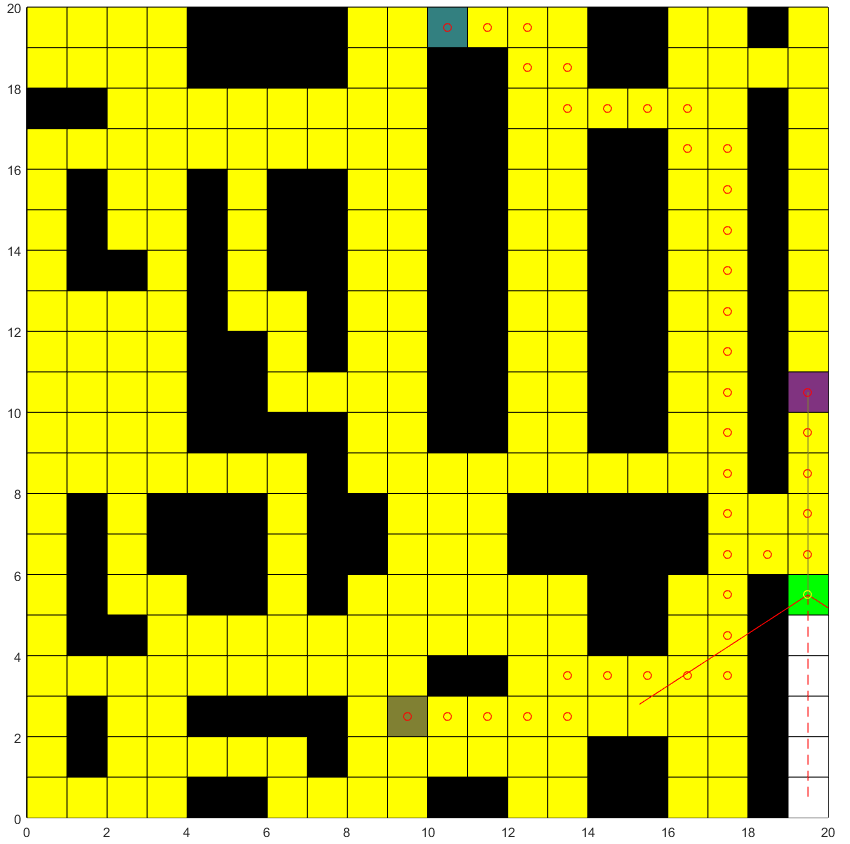}}
\label{fig:simulation_results}
\caption{Representative simulation steps (best viewed in color).}
\vspace{-0.7cm}
\end{figure}

\begin{figure}[!t]
\centering
\includegraphics[width=.6\textwidth]{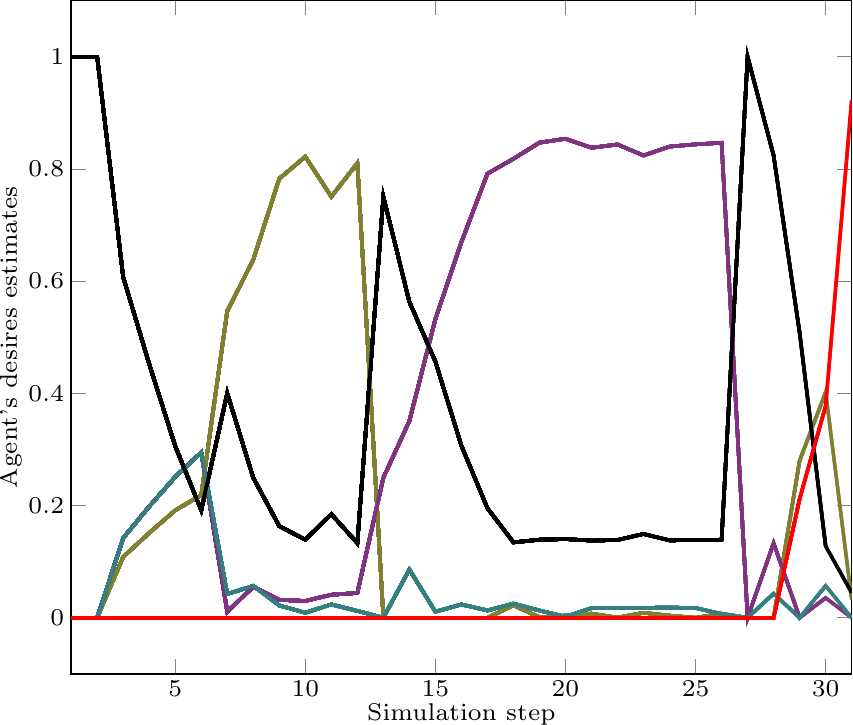}
\caption{Hidden state (desires) probabilities. Probabilities of the goal states are colored according to the goal tile's color. The unknown goal state probability is colored black and irrational agent state probability is colored red.}
\label{fig:probabilities_10}
\vspace{-0.5cm}
\end{figure}

In the previous sections, we have introduced the MDP and HMM frameworks for modeling human action validation and
intention recognition. We have obtained the model parameters empirically and conducted multiple
simulations evaluating the proposed human intention recognition algorithm. The proposed algorithm is tested in a 
 scenario, where the most important simulation steps are shown in Fig.~\ref{fig:simulation_results}
and the corresponding desire estimates are shown in Fig.~\ref{fig:probabilities_10}. The starting position is $(x_1,
y_1, \theta_1)=(5, 5, \frac{3\pi}{2})$. The agent behaves consistently with all the hypotheses and proceeds to the state
$(x_6, y_6, \theta_6)=(8, 5, 0)$. Because of the mentioned hypothesis consistency, the desire estimates for all of the
goal states increase. The actions from simulation step 7 to step 12 are consistent only with the hypothesis $H_3$ which
manifests as the steep rise of the $P(G_3)$ and fall of probabilities related to other goal hypotheses. In the step 13,
action ``Stay'' is the only action consistent with the hypothesis $H_3$ and because the agent chooses the action ``Right'',
the $P(G_3)$ instantly falls towards the zero and $P(G_?)$ and $P(G_2)$ rise. While it might seem obvious that the agent
now actually wants to go to the Goal 2, it has previously chosen actions inconsistent with that hypothesis and the model
initially gives greater probability value to the desire $G_?$ than to $G_2$. Next few steps are consistent with the
hypothesis $H_2$ and the $P(G_2)$ rises until the simulation step 18, when it enters steady state of approximately 0.85.
The goal desires will never obtain value of $1$ because the element $B_{4}'$ is never zero, thus allowing agent's
change of mind. In the state $(x_{25}, y_{25}, \theta_{25})=(16, 7, \frac{\pi}{4})$ agent can decide to go to the Goal 1 or
Goal 2. However, it chooses to take the turn towards the dead end in the simulation step 31. The proposed model recognizes
that this behavior is inconsistent with all of the hypotheses and the $P(G_x)$ steeply rises to value slightly smaller
than 1, declaring the agent irrational.

\section{Conclusion}
In this paper we have proposed a feasible human intention recognition algorithm.
Our goal was to estimate the intention of a human worker, i.e., agent, inside of a robotized warehouse, where we assumed
that the agent's position and orientation are known, as well as the potential goals.
The proposed approach is based on the Markov decision process, where first we run offline the value iteration algorithm
for known agent goals and discretized possible agent states.
The resulting state values are then used within the hidden Markov model framework to generate observations and estimate
the final probabilities of agent's intentions.
Simulations have been carried out within a simulated 2D warehouse with three potential goals, modeling a situation where
the human worker should need to enter the robotized part of the warehouse and pick an item from a rack. 
Results suggest that the proposed framework predicts human warehouse worker’s desires in an intuitive manner and within
reasonable expectations.

\section*{Acknowledgement}
This work has been supported from the European Union's Horizon 2020 research and innovation programme under grant
agreement No 688117 ``Safe human-robot interaction in logistic applications for highly flexible warehouses (SafeLog)'' and
has been carried out within the activities of the Centre of Research Excellence for Data Science and Cooperative Systems
supported by the Ministry of Science, Education and Sports of the Republic of Croatia.

\bibliography{literature,literature2}
\bibliographystyle{unsrt}
\end{document}